\documentclass[conference]{IEEEtran}
\IEEEoverridecommandlockouts
\usepackage{cite}
\usepackage{amsmath,amssymb,amsfonts}
\usepackage{graphicx}
\usepackage{textcomp}
\usepackage{xcolor}
\usepackage{glossaries}
\usepackage{algorithm}
\usepackage{algorithmicx}
\usepackage{algpseudo code}
\usepackage{syntax}
\usepackage{subfigure}
\usepackage{multicol}
\usepackage[T1]{fontenc}

\makeglossaries

\begin{document}
\newacronym{ge}{GE}{Grammatical Evolution}
\newacronym{gp}{GP}{Genetic Programming}
\newacronym{pge}{PGE}{Probabilistic Grammatical Evolution}
\newacronym{sge}{SGE}{Structured Grammatical Evolution}
\newacronym{dsge}{DSGE}{Dynamic Structured Grammatical Evolution}
\newacronym{ea}{EA}{Evolutionary Algorithm}
\newacronym{cfg}{CFG}{Context-Free Grammar}
\newacronym{pcfg}{PCFG}{Probabilistic Context-Free Grammar}
\newacronym{eda}{EDA}{Estimation Distribution Algorithm}
\newacronym{copge}{Co-PGE}{Co-evolutionary Probabilistic Grammatical Evolution}
\newacronym{pmbge}{PMBGE}{Probabilistic Model Building Grammatical Evolution}
\newacronym{cdt}{CDT}{Conditional Dependency Tree}
\newacronym{nt}{NT}{Non-terminal}
\newacronym{pige}{$\pi$GE}{Position Independent Grammatical Evolution}
\newacronym{cfggp}{CFG-GP}{Context-Free Grammar Genetic Programming}
\newacronym{rrse}{RRSE}{Root Relative Squared Error}
\newacronym{ec}{EC}{Evolutionary Computation}
\newacronym{adf}{ADF}{Automatically Defined Functions}
\newacronym{psge}{PSGE}{Probabilistic Structured Grammatical Evolution}
\newacronym{tag}{TAG}{Tree-Adjunct Grammar}
\newacronym{tage}{TAGE}{Tree-Adjunct Grammatical Evolution}
\newacronym{pigrow}{PI Grow}{Position Independent Grow}
\newacronym{ptc2}{PTC2}{Probabilistic Tree Creation 2}
\newacronym{mi}{MI}{Mutation Innovation}
\newacronym{ci}{CI}{Crossover Innovation}
\newacronym{ai}{AI}{Artificial Intelligence}
\newacronym{copsge}{Co-PSGE}{Co-evolutionary Probabilistic Structured Grammatical Evolution}
\newacronym{mgga}{mGGA}{meta-Grammar Genetic Algorithm}
\newacronym{ge2}{(GE)$^2$}{Grammatical Evolution by Grammatical Evolution}

\title{Probabilistic Structured Grammatical Evolution}

\author{\IEEEauthorblockN{Jessica Mégane}
\textit{University of Coimbra}\\
\IEEEauthorblockA{\textit{CISUC}\\
\textit{Dept. of Informatics Engineering} \\
Coimbra, Portugal \\
jessicac@dei.uc.pt}
\and
\IEEEauthorblockN{Nuno Lourenço}
\textit{University of Coimbra}\\
\IEEEauthorblockA{\textit{CISUC}\\
\textit{Dept. of Informatics Engineering} \\
Coimbra, Portugal \\
naml@dei.uc.pt}
\and
\IEEEauthorblockN{Penousal Machado}
\textit{University of Coimbra}\\
\IEEEauthorblockA{\textit{CISUC}\\
\textit{Dept. of Informatics Engineering}\\
Coimbra, Portugal \\
machado@dei.uc.pt}
% \and
% \and
% \and
}
% \author{\IEEEauthorblockN{Anonymous Authors}}

\maketitle

\begin{abstract}
The grammars used in grammar-based \gls{gp} methods have a significant impact on the quality of the solutions generated since they define the search space by restricting the solutions to its syntax. In this work, we propose \gls{psge}, a new approach that combines the \gls{sge} and \gls{pge} representation variants and mapping mechanisms. The genotype is a set of dynamic lists, one for each non-terminal in the grammar, with each element of the list representing a probability used to select the next \gls{pcfg} derivation rule.
\gls{psge} statistically outperformed \gls{ge} on all six benchmark problems studied. In comparison to \gls{pge}, \gls{psge} outperformed 4 of the 6 problems analyzed.
\end{abstract}


\glsresetall

\section{Introduction}
\label{sec:introduction}

\glspl{ea} are metaheuristic algorithms driven by an objective function that follow a trial-and-error approach to problem-solving. 
Inspired by the principles of natural selection and genetics, these algorithms evolve a population of individuals towards better solutions, using an objective function, over several generations.
%\glspl{ea} is the name given to a set of stochastic search procedures that are loosely inspired by the principles of natural selection and genetics. These methods iteratively improve a set of candidate solutions, usually referred to as the population, guided by an objective function.
The quality of solutions improves by selecting the most promising ones (taking into account the objective function), and applying stochastic variations using operators similar to mutations and recombinations that take place in biological systems, where individuals with higher fitness are more likely to survive and reproduce.

\gls{gp} \cite{Koza1994} is a branch of \glspl{ea} in which individuals are represented as computer programs that evolve to solve problems, without the need to program the solution explicitly.
\gls{ge} \cite{ONeill2003,Ryan1998,handbookge} is a \gls{gp} approach that uses a \gls{cfg} to constrain the search space of possible solutions. The individuals are represented by a list of integers (i.e., genotype), where each value (i.e., codon) is used to choose a production rule of the grammar until it forms a solution to the problem (i.e., phenotype).

Despite being one of the most widely used \gls{gp} variants, \gls{ge} is not exempt from having some issues \cite{Rothlauf2003,Rothlauf2006}. \gls{ge} has low locality, which means that small changes in the genotype often cause large changes in the phenotype, causing exploitation to be replaced for exploration, which can lead to a behaviour similar to random search \cite{Whigham2015}. \gls{ge} also suffers from high redundancy, which means that often changes in the genotype do not cause changes in the phenotype \cite{Rothlauf2003}.
To overcome these issues, numerous methods have been proposed in the literature. Most of these methods perform changes in grammars \cite{Harper2010,Nicolau2018,nicolau2004,ONeillRyan2004}, representation of individuals \cite{Kim2015,KIM2016,Lourenco2016,Megane2021,ONeill2004,Ryan2002} or population initialization \cite{Fagan2016,Luke2000,Murphy2012,Nicolau2017,ryanazad2003}.

% --------------------------------- NOVO TEXTO ----------------------
In this paper we propose a new method called \gls{psge} that results from the combination of the representation of \gls{sge} and the mapping mechanism of \gls{pge}. The main motivation for this work rises from the interest of creating a method that inherits the main advantages of the representation used by \gls{sge}, namely the high locality and low redundancy \cite{Medvet2017,Loureno2016}, with the probabilistic mapping form \gls{pge} that is able to guide the evolutionary process towards better solutions by iteratively adjusting the biases of the grammar.

Identically to \gls{sge}, the genotype in \gls{psge} is a set of dynamic lists of real values, with a list for each non-terminal of the grammar. Each element of the list (i.e., codon) represents the probability of choosing a production rule from the grammar, which are updated based on the phenotype of the best individual at the end of each generation, using the same mapping mechanism proposed by \gls{pge}.
% ---------------------------- NOVO TEXTO ---------------------------
\gls{psge} is compared with \gls{ge}, \gls{pge} and \gls{sge} on six different benchmark problems and the results showed that \gls{psge} version is better than \gls{ge} and \gls{pge}.
%is statistically better than \gls{ge} on all problems and that \gls{psge} outperformed \gls{pge} with statistical differences in 4 of the 6 problems analyzed. In terms of relative performance, \gls{psge} and \gls{sge} had similar results.

The remainder of this work is structured as follows: Section \ref{sec:ge} presents the background necessary to understand the work presented, introducing \gls{ge} as well as related work.
Section \ref{sec:psge} present the proposed method, detailing the representation and mapping method used. Section \ref{sec:val} details the experimentation framework used and Section \ref{sec:results} the experimental results regarding performance. Section \ref{chap:conclusion} gathers the main conclusions and provides some insights regarding future work.

\section{Grammatical Evolution}
\label{sec:ge}

% \gls{ge} \cite{ONeill2003,Ryan1998,handbookge} is a grammar-based \gls{ea} in which individuals are represented as a list of integers. The elements of the list are mapped creating the phenotype of the individual (i.e., solution to the problem), using production rules defined by a \gls{cfg}.
\gls{ge} \cite{ONeill2003,Ryan1998,handbookge} is a grammar-based \gls{gp} approach used to evolve programs. Each individual in the population is represented by a list of integers, called genotype, and each value is randomly generated in the interval $[0, 255]$. The elements of the list are used in the mapping process to create the phenotype, that is the solution of the problem, using the production rules defined by a \gls{cfg}.
A grammar is a tuple $G = (NT,T,S,P)$ where $NT$ and $T$ represent the non-empty set of \textit{Non-Terminal (NT)} and \textit{Terminal (T)} symbols, $S$ is an element of $NT$ in which the derivation sequences start, called the axiom, and $P$ is the set of production rules. The rules in $P$ are in the form $A ::= \alpha$, with $A \in NT$ and $\alpha \in (NT \cup T)^+$. The $NT$ and $T$ sets are disjoint. Each grammar defines a language $L(G)= \{ w:\, S\overset{*} {\Rightarrow} w,\, w \in T^*\}$, that is the set of all sequences of terminal symbols that can be derived from the axiom. The symbol $*$ represents the unary operator Kleene star.

\begin{figure}[htbp]
\centering
		\scalebox{0.65}{\fbox{%
				\parbox{0.1\textwidth}{%
					\begin{align*}
						{<}\text{expr}{>} ::=  & \, {<}\text{expr}{>}{<}\text{op}{>}{<}\text{expr}{>} \\
						& | \, {<}\text{var}{>} \\
						{<}\text{op}{>} ::= & \,+ \, \\
						& | \, - \\
						& | \, * \\
						& | \, / \\
						{<}\text{var}{>} ::= &  \, \text{x} \, \\ 
						& | \, \text{y} \,  \\
						& | \, \text{1.0} \,
		\end{align*}}}}
		\caption{\label{gemapgram}Example of a CFG.}
	\end{figure}
%The genotype-phenotype mapping begins with the axiom of the grammar, and the expansion is always made from the leftmost non-terminal. Each codon is mapped into a production rule by applying the modulo operator ($mod$) between the codon and the number of expansion options of the non-terminal to expand.
The mapping from genotype to phenotype is done starting from the axiom of the grammar and expanding the leftmost non-terminal. The codons of the genotype are used to choose which production rule to expand by applying the modulo operator ($mod$) between the codon and the number of derivation rules of the respective non-terminal.

\begin{figure}[htbp]
\centering
		\scalebox{0.65}{\fbox{%
				\parbox{0.1\textwidth}{%
					\begin{align*}
						\textbf{Gen} & \textbf{otype}\\
						{[}\text{34, 13, 9, 151, 95, 221, } & \text{23, 98, 145, 42, 153}{]}\\
						{<}\text{expr}{>} \rightarrow  & \, {<}\text{expr}{>}{<}\text{op}{>}{<}\text{expr}{>} & 34 mod 2 = 0 \\
						{<}\text{expr}{>}{<}\text{op}{>}{<}\text{expr}{>} \rightarrow & \, {<}\text{var}{>}{<}\text{op}{>}{<}\text{expr}{>} & 13 mod 2 = 1 \\
						{<}\text{var}{>}{<}\text{op}{>}{<}\text{expr}{>} \rightarrow & \, \text{x} \, {<}\text{op}{>}{<}\text{expr}{>} & 9 mod 3 = 0\\
						\text{x}{<}\text{op}{>}{<}\text{expr}{>} \rightarrow & \, \text{x} \, \text{/} \, {<}\text{expr}{>} & 151 mod 4 = 3\\
						\text{x} \, \text{/} \, {<}\text{expr}{>} \rightarrow & \, \text{x} \, \text{/} \, {<}\text{var}{>} & 95 mod 2 = 1 \\
						\text{x} \, \text{/} \, {<}\text{var}{>} \rightarrow & \, \text{x} \, \text{/} \, \text{1.0} & 221 mod 3 = 2 \\
						\textbf{Phenotype: } & x / 1.0
		\end{align*}}}}
	\caption{\label{fig:gemapping}Example of the genotype-phenotype mapping of GE.}
\end{figure}

An example of the genotype-phenotype process is shown in Fig. \ref{fig:gemapping}, using the example of grammar presented in Fig. \ref{gemapgram}.
%The genotype is composed of integers values (i.e., codons) randomly generated in the interval $[0, 255]$.
The mapping begins with the axiom of the grammar, $<$expr$>$, which has two expansion alternatives, and the first unused codon of the genotype, $34$. By applying the modulo operator between the codon and the number of production rules, $34 mod(2) = 0$, we obtain the index of the rule to be expanded which is $<$expr$><$op$><$expr$>$.
% This process is performed until there are no more non-terminal symbols to expand or there are no more integers to read from the genotype.
% In this last case and if we still have non-terminals to expand, a wrapping mechanism can be used, where the genotype will be re-used until it generates a valid individual or the predefined number of wraps is over. If after all the wraps we still have not mapped all the non-terminals, the mapping process stops, and the individual will be considered invalid. The phenotype of each individual is evaluated with the fitness function and then the population goes through the selection mechanisms.
This procedure is repeated until there are no more non-terminal symbols to expand or numbers from the genotype to read.
In the last case, if we still have non-terminals to expand, we can employ a wrapping technique, in which the genotype is reused until it yields a valid individual or the predefined number of wraps is reached. If we haven't mapped all of the non-terminals after all of the wraps, the mapping procedure will stop and the individual will be considered invalid. The fitness function is used to evaluate each individual's phenotype, and then the population is subjected to selection procedures.

% ----------------- HERE -------------------------

\subsection{Related Work}
\label{sec:related}
\gls{ge} is one of the most popular \gls{gp} variants, and it has undergone various modifications over the years to address some of its major criticisms, namely high redundancy and low locality. Low locality refers to small genotype changes that result in large phenotypic changes, whereas high redundancy refers to multiple genotypes corresponding to the same phenotype.
The majority of these proposed solutions include changes to grammars \cite{Harper2010,nicolau2004,Nicolau2018,ONeillRyan2004}, individual representation \cite{Kim2015,KIM2016,Loureno2018,Megane2021,ONeill2004,Ryan2002}, or population initialization \cite{Fagan2016,Luke2000,Murphy2012,Nicolau2017,ryanazad2003}.

% ---------------------------------------- DEIXO PARA O FIM
\gls{sge} \cite{Loureno2018} addresses \gls{ge}'s locality and redundancy issues while achieving better performance results \cite{Lourenco2016}. 
The genotype in \gls{sge} is a set of dynamic lists of ordered integers, one list for each non-terminal of the grammar. Each value in the list represents which production rule to select from the non-terminal.
Different grammar-based \gls{gp} approaches were compared in \cite{Loureno2017}, and the authors demonstrated that \gls{sge} outperformed several grammar-based \gls{gp} representations in some problems.

% Pi GE
% \gls{pige} \cite{ONeill2004} is a method that introduces a different representation and mapping mechanism, in which the order of expansion of the non-terminals is determined by the genotype of the individual, removing the positional dependency that exists in \gls{ge}. The genotype of the individuals is composed of two values (\textit{nont}, \textit{rule}), where \textit{nont} used to select the next non-terminal to be expanded, and \textit{rule} selects which rule to derive from that non-terminal. This method proved to be better than \gls{ge} on several problems, showing statistical differences \cite{Fagan2010}.

\gls{pige} \cite{ONeill2004} is a method that uses a new representation and mapping mechanism in which the genotype of the individual determines the order of expansion of the non-terminals, reducing the positional dependency that exists in \gls{ge}. Individual genotypes are made of tuples of two values (\textit{nont}, \textit{rule}), with \textit{nont} determining which non-terminal to expand next and \textit{rule} determining the rule to derive from that non-terminal. On several problems, this technique outperformed \gls{ge}, with statistical differences.

% Chorus
% Chorus \cite{Ryan2002} is another method in which there is positional independence, with each gene specifically encoding one production of the grammar. However, this approach has not been shown to be better than the \gls{ge} standard.
Chorus \cite{Ryan2002} is another method that allows for positional independence, with each gene encoding only one production of the grammar. This strategy, however, has not been proved to be superior than the \gls{ge} standard.

% % grammar design
% The design of the grammars is another aspect that has had some attraction for researchers, since they define the search space, and so the choice of grammar can influence the speed of convergence to the best solution \cite{Nicolau2018}. 
% Some studies have been conducted to analyze the performance of \gls{ge} with different types of grammars, such as the use of recursively balanced grammars \cite{Harper2010,Nicolau2018} and the reduction of non-terminal symbols \cite{nicolau2004,Nicolau2018}.

The design of the grammars is another aspect that has had some attraction for researchers since they define the search space, and thus the choice of grammar can affect the speed of convergence to the best solution \cite{Nicolau2018}.
Some research has been done to study the performance of \gls{ge} with various types of grammars, such as the use of recursively balanced grammars \cite{Harper2010,Nicolau2018} and the reduction of non-terminal symbols \cite{nicolau2004,Nicolau2018}.

Harper et al. \cite{Harper2010} demonstrated that the grammar used at the start of the evolutionary process can have a significant impact on the solutions, such as generating a large number of invalid individuals when using recursive grammars. It has also been demonstrated that when a balanced grammar is used, there is more variety in the size of solutions.

Nicolau et al. \cite{Nicolau2018} tested \gls{ge} with different grammars, which included balanced grammars, grammars with corrected biases, and grammars with unlinked productions.
The experimental tests using a recursively balanced grammar, in which there is a non-recursive production for every recursive one, yielded better results than the original grammar. However it resulted in a larger number of individuals consisting of a non-terminal symbol.
Nicolau \cite{nicolau2004} proposed a method for reducing the number of non-terminal symbols, replacing them with their productions, which, while showing a slight improvement in performance, has the disadvantage of producing complex grammars that are difficult to read.

Another area of research has been the evolution of the grammar throughout the evolutionary process \cite{ONeillRyan2004,Megane2021,Kim2015,KIM2016}.

% \gls{ge} by ge
\gls{ge2} \cite{ONeillRyan2004} is a method where the grammar and genetic code co-evolve. The method employs two distinct grammars: universal grammar and solution grammar. The structure of the solution grammar, which is used to map the individuals, is dictated by the universal grammar. This method has been shown to be effective in developing biases toward non-terminal symbols. Later, it was implemented into a new algorithm, \gls{mgga} \cite{ONeill2005}, which resulted in improved performance.

% is an approach in which there is co-evolution of grammar and genetic code. The method uses two distinct grammars, the universal grammar and the solution grammar. The universal grammar dictates the structure of the solution grammar, that is used to map the individuals. This method has shown to be effective in evolving biases towards some non-terminal symbols. Later, was implemented into a new algorithm, \gls{mgga} \cite{ONeill2005}, which obtained performance improvements. 

\gls{pge} \cite{Megane2021} is a recent variant of \gls{ge} in which the individuals are mapped using a \gls{pcfg} and the genotype is a list of real numbers. A \gls{pcfg} is a quintuple $PG = (NT,T,S,P,Probs)$ where $NT$ and $T$ represent the non-empty set of \textit{Non-Terminal (NT)} and \textit{Terminal (T)} symbols, $S$ is an element of $NT$ called the axiom, $P$ is the set of production rules, and $Probs$ is a set of probabilities associated with each production rule. The mapping begins with the leftmost non-terminal, and the rule whose probability interval includes the codon is chosen for each non-terminal to be expanded. The \gls{pcfg} probabilities are updated at the end of each generation based on the expansion rules used to create the best individual of the current generation alternating with the best individual overall. \gls{pge} proved to be superior than \gls{ge} with statistical differences in the two problems studied.

Kim et al. \cite{Kim2015} proposed \gls{pmbge}, in which the mapping is based on a \gls{pcfg} and the probabilistic technique \gls{eda}, which also replaces the mutation and crossover operators. The probabilities of the grammar are changed every generation based on the frequency of the rules expanded by the best individuals. This technique generates a new population from the new grammar at each generation. When compared to \gls{ge}, the proposed approach performed similarly.
Later, Kim et al.\cite{KIM2016} adapted \gls{cdt} to the mechanism of updating the grammar, creating cd\gls{pmbge} which takes into account the dependencies between production rules. The results revealed no statistical differences between \gls{ge} and the proposed approach.
This method outperformed \gls{ge} with statistical differences in two of the four problems analysed.

\section{Probabilistic Structured Grammatical Evolution}
\label{sec:psge}

In this work we propose \gls{psge} \footnote{The implementation of PSGE is available at:\\ https://github.com/jessicamegane/psge} in which the representation of individuals and the mapping mechanism is a combination of the approaches followed by \gls{sge} and \gls{pge}.
The mapping resorts to a \gls{pcfg} to choose the derivation rules. At the end of each generation the probabilities of the grammar are updated according to the production rules expanded by the best individual of the current generation or the best individual overall.

The individuals are represented by a set of dynamic lists, with each list being associated with a non-terminal of the grammar.
The lists contain an ordered sequence of real numbers, with each codon corresponding to the probability of choosing a production rule.

The pseudo-code of the initialization of individuals in \gls{psge} is presented in Alg. \ref{psgeindividual}. The algorithm takes as parameters the genotype (which starts with an empty list for each non-terminal), the non-terminal symbol to expand (the axiom of the grammar in the first iteration), the current depth (which starts at 0), the maximum depth limit and the \gls{pcfg}.
The function is recursive and ends when the genotype belongs to a valid individual. At each iteration a random value between 0 and 1 is generated and added to the list of the non-terminal to be expanded (Alg. \ref{psgeindividual}, lines 2-3). To determine which rule to expand next based on the new codon, the mapping process is simulated (Alg. \ref{psgeindividual}, line 4).

\begin{algorithm}[htbp]
  %\scriptsize
  \caption{Random candidate solution of PSGE}
	\label{psgeindividual}
	\begin{algorithmic}[1]
		\Procedure{createIndividual}{genotype, symb, depth, max\_depth, pcfg}
		\State codon = $random$(0,1)
		\State genotype[symb].append(codon)
		%\If{$current\_depth > max\_depth$ \textbf{and} $is\_recursive(symb)$}
		%\State selected\_rule = generate\_terminal\_expansion(symb, codon, pcfg)
		%\Else
		\State selected\_rule = $generate\_expansion$(symb, codon, pcfg, depth, max\_depth)
		%\EndIf		
		\State expansion\_symbols = pcfg[symb][selected\_rule]
		\For{sym \textbf{in} expansion\_symbols}
		\If{\textbf{not} $is\_terminal$(sym)}
		\State $createIndividual$(genotype, symb, depth + 1, max\_depth, pcfg)
		\EndIf
		\EndFor
		\EndProcedure
	\end{algorithmic}
\end{algorithm}

The approach for genotype-phenotype mapping is described in Alg. \ref{psgemapping}. The genotype, a counter called $positions\_to\_map$ (which is initially empty and is used to store the genotype position of each non-terminal list at the current iteration), the symbol to expand (which starts in the axiom), the current depth, the maximum depth limit, and the grammar are all passed as arguments to the algorithm.
If more codons are required to construct a valid individual during mapping, they will be created at random and added to the genotype (Alg. \ref{psgemapping}, lines 3-6). One of the benefits of this representation is that with the depth limit, it is feasible to add productions as needed without the risk of bloat (a significant increase in the size of the solutions \cite{Eiben2015}), ensuring that valid individuals are always created.

\begin{algorithm}[htbp]
	\caption{Genotype-Phenotype mapping of PSGE}
	\label{psgemapping}
	\begin{algorithmic}[1]
		\Procedure{mapping}{genotype, positions\_to\_map, symb, depth, max\_depth, pcfg}
		\State phenotype = ""
		% add novo codao caso seja necessario
		\If{positions\_to\_map[symb] $>=$ $len$(genotype[symb])}
		\State codon = $random$(0,1)
		\State genotype[symb].append(codon)
		\EndIf
		% remapping
		\State codon = genotype[symb][positions\_to\_map[symb]]
		\State selected\_rule = $generate\_expansion$(symb, codon, pcfg, current\_depth, max\_depth)
		\State expansion = pcfg[symb][selected\_rule]
		\State positions\_to\_map[symb] += 1
		\For{sym \textbf{in} expansion}
		\If{$is\_terminal$(sym)}
		\State phenotype += sym
		\Else
		\State phenotype += $mapping$(genotype, positions\_to\_map, sym, depth + 1, max\_depth, pcfg)
		\EndIf
		\EndFor
		\State \textbf{return} phenotype
		\EndProcedure
	\end{algorithmic}
\end{algorithm}

To choose the derivation rule to expand, we use the function that is described in Algorithm \ref{psgerule}. The function receives as parameters the non-terminal symbol to be expanded, the codon, the grammar, the current depth and the maximum depth established. The mechanism for choosing the derivation rule is based on that of \gls{pge} \cite{Megane2021}, in which the rule whose probability interval incorporates the value of the codon is chosen (Alg. \ref{psgerule}, lines 15-21), however for \gls{psge} when the depth limit is reached, only non-recursive production rules are considered (Alg. \ref{psgerule}, lines 4-13). In this case, the probabilities of the non-recursive rules are adjusted proportionally so that their sum is 1. The codon value is compared with the new probability values, and the rule whose probability range includes the codon value is chosen.

% The process of choosing a derivation rule from a codon of the genotype using a \gls{pcfg} is similar to that used by \gls{pge} \cite{Megane2021} (Alg. \ref{psgerule}, lines 15-21), except that there is a distinction when the maximum depth limit is exceeded, in which only non-recursive productions can be chosen (Alg. \ref{psgerule}, lines 4-13). The function receives as parameters the non-terminal symbol to be expanded, the codon, the grammar, the current depth and the maximum depth established.

% When the defined maximum tree depth limit is exceeded, the algorithm considers only non-recursive rules, and adjusts the probabilities of each of them, so that the sum is 1. To accomplish this, we first sum the value of the current probabilities of the non-recursive rules, which is used to perform the adjustment. Using the new probabilities, the production rule is chosen with the normal procedure: It is verified whether the codon belongs to the probability range of each production rule of the non-terminal to be expanded and when this condition is verified, the rule is chosen.

\begin{algorithm}[htbp]
	\caption{PSGE function to select an expansion rule}
	\label{psgerule}
	\begin{algorithmic}[1]
		\Procedure{generate\_expansion}{symb, codon, pcfg, depth, max\_depth}
		\State cum\_prob = 0.0
		\If{depth $\geq$ max\_depth}
		\State nr\_prods = $get\_non\_recursive\_prods$(pcfg[symb])
		\State total\_nr\_prods = $sum$(nr\_prods.$getProb$())
		\For{prod \textbf{in} non\_recursive\_prods}
		\State new\_prob = prod.$getProb$() / total\_nr\_prods
		\State cum\_prob = cum\_prob + new\_prob
		\If{codon $\leq$ cum\_prob}
		\State selected\_rule = prod 
		\State \textbf{break}
		\EndIf
		\EndFor
		\Else
		\For{prod \textbf{in} pcfg[symb]}
		\State cum\_prob = cum\_prob + prod.$getProb()$
		\If{codon $\leq$ cum\_prob}
		\State selected\_rule = prod
		\State \textbf{break}
		\EndIf
		\EndFor
		\EndIf
		\State \textbf{return} selected\_rule
		\EndProcedure
	\end{algorithmic}
\end{algorithm}

% ------------------------ HERE------------------------------------
% The mapping process in \gls{psge} is illustrated in Fig. \ref{fig:mappingpsge} using the example of \gls{pcfg} shown on Fig. \ref{psgemapgram}. 

\begin{figure}[htbp]
\centering
	\noindent\scalebox{0.65}{\fbox{%
			\parbox{0.1\textwidth}{%
				\begin{align*}
				\textbf{PCFG} & & \textbf{Prob.} \quad      \\
					{<}\text{expr}{>} ::=  & \, {<}\text{expr}{>}{<}\text{op}{>}{<}\text{expr}{>}  &  [0.00;0.37]\\
					& | \, {<}\text{var}{>} & ]0.37;1.00]\\
					{<}\text{op}{>} ::= & \,+ \,  & [0.00;0.22]\\
					& | \, -  & ]0.22;0.39]\\
					& | \, *  & ]0.39;0.68]\\
					& | \, /   & ]0.68;1.00]\\
					{<}\text{var}{>} ::= &  \, \text{x}  & [0.00;0.41]\\ 
					& | \, \text{y}  & ]0.41;0.67] \\
					& | \, \text{1.0}  & ]0.67;1.00]
	\end{align*}}}}
\caption{\label{psgemapgram}PCFG example. Prob. represents the range of values covered by each production rule, and the size of the range corresponds to the probability of that production rule being chosen.}
\end{figure}

An example of an individual's mapping process is depicted in Fig. \ref{fig:mappingpsge}, using the grammar in Fig. \ref{psgemapgram}. 
The process begins by expanding the axiom of the grammar, $<$expr$>$, using the first codon in the list of the respective non-terminal of the genotype, in this case $0.19$.
The non-terminal $<$expr$>$ presents two derivation rules, with different probabilities of being chosen. The $0.19$ codon is included in the range of probabilities of the first rule, $<$expr$><$op$><$exp$>$, therefore expansion is made for that rule.
The derivation is always done from the leftmost non-terminal, so the next non-terminal to expand is $<$expr$>$. The next available codon in the list of the non-terminal $<$expr$>$ is $0.46$, which falls within the probability range of the second expansion rule, $<$var$>$.
The $<$var$>$ will be the next to expand and has three derivation rules. $0.32$ is the first codon available in the list of $<var>$ of the genotype, and falls within the range of probabilities covered by the first production rule, $x$, which is a terminal symbol, so $<$op$>$ is the next symbol to expand. The procedure is repeated until a valid individual is formed.

\begin{figure}[htbp]
\centering

\scalebox{0.65}{\fbox{%
		\parbox{0.1\textwidth}{%
			\begin{align*}
				\textbf{Genot} & \textbf{ype}\\ 
				%\framebox[3cm]{<expr>}\framebox[2cm]{<op>}&\framebox[2.5cm]{<var>} \\
				\framebox[2.8cm]{<expr>}\framebox[2cm]{<op>}&\framebox[2.5cm]{<var>} \\
				\framebox[2.8cm]{[0.19,0.46,0.87]}\framebox[2cm]{[0.27]}&\framebox[2.5cm]{[0.32, 0.64]} \\
				{<}\text{expr}{>} \rightarrow  & \, {<}\text{expr}{>}{<}\text{op}{>}{<}\text{expr}{>} & (0.19) \\
				{<}\text{expr}{>}{<}\text{op}{>}{<}\text{expr}{>} \rightarrow & \, {<}\text{var}{>}{<}\text{op}{>}{<}\text{expr}{>} & (0.46)\\
				{<}\text{var}{>}{<}\text{op}{>}{<}\text{expr}{>} \rightarrow & \, \text{x} \, {<}\text{op}{>}{<}\text{expr}{>} & (0.32)\\
				\text{x}{<}\text{op}{>}{<}\text{expr}{>} \rightarrow & \, \text{x} \, \text{-} \, {<}\text{expr}{>} & (0.27)\\
				\text{x} \, \text{-} \, {<}\text{expr}{>} \rightarrow & \, \text{x} \, \text{-} \, {<}\text{var}{>} & (0.87) \\
				\text{x} \, \text{-} \, {<}\text{var}{>} \rightarrow & \, \text{x} \, \text{-} \, \text{y} & (0.64) \\
				\textbf{Phenotype: } & x - y
\end{align*}}}}

	\caption{Example of the genotype-phenotype mapping of PSGE with a PCFG.}
	\label{fig:mappingpsge}
\end{figure}

% ------------------ FEITO ------------------------

At the end of each generation the best individual overall and the best individual from the current generation are used alternately to update the \gls{pcfg} probabilities, using the same mechanism proposed for \gls{pge} \cite{Megane2021}. All individuals are re-mapped to update the phenotype according to the new updated grammar. 

For each production \textit{i} of each non-terminal \textit{j} we have a \textit{counter} with the number of times that each production was chosen, and the probability (\textit{prob}) of the \gls{pcfg} of choosing that production. If the counter is greater than zero, that is, the production rule was used to map the individual, we use \eqref{eq:pgecounterpos}. If the counter is zero, that is, the production rule has not been used by the individual, we use \eqref{eq:pgecounterzero}.
The learning factor is represented by $\lambda$, with $\lambda \in [0, 1]$, and is used to make the transitions on the search space smoother. 

\begin{equation}
	\label{eq:pgecounterpos}
	prob_i = min(prob_i + \lambda * \frac{counter_i}{\sum_{k=1}^{j} counter_k}, 1.0)
\end{equation}

\begin{equation}
	\label{eq:pgecounterzero}
	prob_i = prob_i - \lambda * prob_i
\end{equation}

% where $\lambda$ is the learning factor, with $\lambda \in [0,1]$, used to make the transitions on the search space smoother.

After updating the probabilities using the equations, these are adjusted until the sum of the probabilities of each production rule of each non-terminal is 1.

\subsection{Variation Operators}
\label{sec:cross}
% ---- genetic operators ----------
Genetic operators are used to introduce genotype-level changes in individuals, such as mutation and crossover.
The mutation operator changes randomly chosen codons, and in \gls{psge} a Gaussian mutation is applied to these codons, keeping the new value in the interval $[0, 1]$. This type of mutation is widely used in the literature and has proven to be an effective method for making small changes in the search space \cite{Beyer2004EvolutionS,hinterdinggaussian}.

\begin{figure}[htbp]
	\centering
\textbf{Genotype before mutation:}		\scalebox{0.7}{\framebox[3cm]{$<expr>$}\framebox[2cm]{$<op>$}\framebox[2.5cm]{$<var>$}} \\
		\scalebox{0.7}{\framebox[2.8cm]{[0.19,\textbf{0.46},0.87]}\framebox[2cm]{[0.27]}\framebox[2.5cm]{[0.32, 0.64]}} \\

		\centering
		\textbf{\\Genotype after mutation:}
		\scalebox{0.7}{\framebox[3cm]{$<expr>$}\framebox[2cm]{$<op>$}\framebox[2.5cm]{$<var>$}} \\
		\scalebox{0.7}{\framebox[2.8cm]{[0.19,\textbf{0.29},0.87]}\framebox[2cm]{[0.27]}\framebox[2.5cm]{[0.32, 0.28]}} \\
		
	\caption{Example of PSGE's mutation on one codon of the genotype.}
	\label{fig:mutationpsge}
\end{figure}
Fig. \ref{fig:mutationpsge} is an example of Gaussian mutation in an individual. Assuming that the second codon of the non-terminal $<expr>$, was randomly selected ($0.46$) and the value generated with a normal distribution of mean 0 and standard deviation 0.50  ($N(0,0.50)$) was $-0.17$, the codon will now assume a value of $0.29$.

\begin{figure}[htbp]
	\centering
		\centering
		\textbf{Parents genotype:}\\
		Parent 1:\\
		\scalebox{0.7}{\framebox[3cm]{$<expr>$}\framebox[2cm]{$<op>$}\framebox[2.5cm]{$<var>$}} \\
		\scalebox{0.7}{\framebox[3cm]{[0.19,0.46,0.87]}\framebox[2cm]{[0.27]}\framebox[2.5cm]{[0.32,0.64]}} \\
		\textbf{ }\\
		Parent 2:\\
		\scalebox{0.7}{\framebox[3cm]{$<expr>$}\framebox[2cm]{$<op>$}\framebox[3cm]{$<var>$}} \\
		\scalebox{0.7}{\framebox[3cm]{[0.02,0.90,0.13]}\framebox[2cm]{[0.48]}\framebox[3cm]{[0.75,0.42,0.56]}} \\
		
		\centering
		\textbf{\\Mask and offspring after crossover:}\\
		Mask:\\
		\scalebox{0.7}{\framebox[2cm]{$<expr>$}\framebox[2cm]{$<op>$}\framebox[2cm]{$<var>$}} \\
		\scalebox{0.7}{\framebox[2cm]{1}\framebox[2cm]{1}\framebox[2cm]{0}} \\
		\textbf{ }\\
		Offspring:\\
		\scalebox{0.7}{\framebox[3cm]{$<expr>$}\framebox[2cm]{$<op>$}\framebox[2.5cm]{$<var>$}} \\
		\scalebox{0.7}{\framebox[3cm]{[0.02,0.90,0.13]}\framebox[2cm]{[0.48]}\framebox[2.5cm]{[0.32,0.64]}} \\

	\caption{Example of PSGE's crossover between two individuals, generating one offspring.}
	\label{fig:xoversge}
\end{figure}

The crossover operator combines the genetic material of two individuals to generate an offspring, and is based on the crossover proposed by \gls{sge} \cite{Loureno2018}. The offspring inherits the list of each non-terminal from one of the parents, and this decision is made based on a randomly generated binary mask. The mask contains a binary value for each list of the genotype (i.e., one for each non-terminal of the grammar). In Fig. \ref{fig:xoversge} are represented an example of crossover, showing the parents, the mask used and the offspring. In the example, the descendant inherited the lists of the non-terminals $<expr>$ and $<op>$ from Parent 2, and the list of the non-terminal $<var>$ from Parent 1. In case two descendants are generated, the other would get the opposite lists.
\section{Experimental Setup}
\label{sec:val}

The performance of \gls{psge} will be carried out following the framework proposed by Whigham et al. \cite{Whigham2015}, examining the evolution of the mean best fitness of each generation over 100 independent runs in six problems of different scopes. The results will be compared with the standard \gls{ge}, \gls{pge}, and \gls{sge}. %TODO: justificar porque é que comparo com o GE, SGE e PGE
Table \ref{table:parameters} presents the parameters used by all the approaches.

\begin{table}[htbp]
\centering
\caption{\label{table:parameters}Parameters used in the experimental analysis for GE, PGE, SGE and PSGE.}
\begin{tabular}{|c|cccc|}
\hline
\multicolumn{1}{|c|}{\textbf{Parameters}} & \multicolumn{1}{c|}{\textbf{GE}} & \multicolumn{1}{c|}{\textbf{PGE}} & \multicolumn{1}{c|}{\textbf{SGE}} & \textbf{PSGE} \\ \hline
Population Size        & \multicolumn{4}{c|}{1000}                                                            \\ \hline
Generations            & \multicolumn{4}{c|}{50}                                                              \\ \hline
Elitism Count          & \multicolumn{4}{c|}{100}                                                             \\ \hline
Mutation Rate          & \multicolumn{4}{c|}{0.05}                                                            \\ \hline
Crossover Rate         & \multicolumn{4}{c|}{0.90}                                                            \\ \hline
Tournament             & \multicolumn{4}{c|}{3}                                                               \\ \hline
Size of Genotype       & \multicolumn{2}{c|}{128}                           & \multicolumn{2}{c|}{-}          \\ \hline
Max Depth              & \multicolumn{2}{c|}{-}                             & \multicolumn{2}{c|}{10}         \\ \hline
\end{tabular}
\end{table}

%, and in order to validate the effectiveness of the proposed methods, the results were compared with those of \gls{ge} and \gls{sge}.

In what concerns the variation operators, \gls{ge} and \gls{pge} use a one point crossover. The mutation of \gls{ge} replaces the selected codons by new ones randomly generated in the interval $[0, 255]$ and in the case of \gls{pge} a float mutation is used, in which the codons are replaced by new ones generated in the interval $[0, 1]$.
The wrapping mechanism was removed from \gls{ge} and \gls{pge}.
Regarding the genetic operators used by \gls{sge} and \gls{psge}, these methods all use the same crossover, which is the one presented in Section \ref{sec:cross}. Regarding the mutation operator, in the case of \gls{sge}, the mutated codon is replaced with a different valid option, while in \gls{psge} a Gaussian mutation with $N(0, 0.50)$ in the codon value is performed.
%Additionally, \gls{pge} and \gls{psge} use a learning factor of $\lambda = 1.0\%$.

A detailed description of the problems and the grammars used can be found in the work done by Lourenço et al. \cite{Loureno2017} and Whigham et al. \cite{Whigham2015}, and were selected on the basis of the recommendations of McDermott et al. \cite{McDermott2012}.
The problems considered were the quartic symbolic regression, pagie symbolic regression \cite{White2012}, the Boston Housing symbolic regression \cite{BostonHousing}, 5-bit even parity, 1-bit Boolean multiplexer and the Santa Fe artificial ant problem \cite{Koza1994}.
The objective functions used to evaluate the individuals consider the minimization of the error. In the case of Symbolic Regression and classification problems the fitness is the \gls{rrse} between the individual's solution and the target on a data set. For the Boolean functions, the error is the number of incorrect predictions, and for the Path finding problem, the fitness is the number of pieces remaining after exceeding the step limit.

\section{Results}
\label{sec:results}
A statistical analysis was done to be able to fairly compare the different methods and support our analysis. Since the populations were independently initialized and the results do not meet the criteria for parametric tests, the Kruskal-Wallis non-parametric test was used to check for meaningful differences between the different methods.
%To better understand and compare the performance of all approaches we performed a statistical analysis. Since the results did not follow any distribution, and the populations were independently initialised, we employed the Kruskal-Wallis non-parametric test to check if there were meaningful differences between the different approaches.
When the methods show differences, we verify in which pairs the differences exist, using the Mann-Whitney \textit{post-hoc} test with the Bonferroni correction. To determine how significant the differences are, the effect size $r$ was calculated. The "\texttildelow" sign was used when there were no significant differences between samples, the "+" sign was used when the effect size was small ($r \leq 0.3$), "++" was used when the effect size was medium ($0.3 < r \leq 0.5$), and "+++" was used when the effect size is large ($r > 0.5$).
%When this happened we used the Mann-Whitney \textit{post-hoc} test with Bonferroni correction to ascertain in which pairs this difference existed.
For all the statistical tests we considered a significance level of $\alpha = 0.05$.

% \begin{table}[htbp]
% 	\centering
% 	\caption{\label{statstable}Results of the Mann-Whitney post-hoc statistical tests. The Bonferroni correction is used considering a significance level of $\alpha$ = 0.05. Values in bold mean that PSGE is statistically better than GE, PGE or SGE.\\}
% 	\begin{tabular}{ccccc}
% 		\hline
% 		p-value              &  \multicolumn{1}{|}{} \quad \multicolumn{1}{c}{PSGE-GE} \quad \multicolumn{1}{|}{} & \quad PSGE-PGE \quad  \multicolumn{1}{|}{} & \quad PSGE-SGE \quad  \multicolumn{1}{}{}\\ \hline
% 		Quartic Polynomial   & \multicolumn{1}{|}{} \textbf{0.000}      & \textbf{0.000}       & 0.299        \\
% 		Pagie Polynomial     &  \multicolumn{1}{|}{} \textbf{0.033}      & 0.579       & 0.013         \\
% 		Boston Housing Train &  \multicolumn{1}{|}{} \textbf{0.000}      & \textbf{0.003}       & 0.001  \\
% 		Boston Housing Test  &  \multicolumn{1}{|}{} \textbf{0.000}      & 0.060       & 0.577         \\
% 		5-bit Parity         &  \multicolumn{1}{|}{} \textbf{0.000}      & \textbf{0.000}      & 0.131         \\
% 		11-bit Multiplexer   &  \multicolumn{1}{|}{} \textbf{0.000}      & \textbf{0.000}       & 0.000        \\
% 		Santa Fe Ant Trail   &  \multicolumn{1}{|}{}\textbf{0.000}      & \textbf{0.000}       & 0.317         \\ \hline
% 	\end{tabular}
% \end{table}

% bh train psge-pge 1.601
\begin{table}[htbp]
\centering
\caption{\label{statstable}Results of the Mann-Whitney post-hoc statistical tests. The Bonferroni correction is used considering a significance level of $\alpha$ = 0.05. Values in bold mean that PSGE is statistically better than GE, PGE or SGE.\\}
    \begin{tabular}{|l|c|c|c|}
        \hline
        \textbf{Problem}              &  \textbf{PSGE-GE}  &  \textbf{PSGE-PGE}    &  \textbf{PSGE-SGE}  \\ \hline
        Quartic Polynomial   & \textbf{0.000} & \textbf{0.003} & 0.299    \\ \hline
        Pagie Polynomial     & \textbf{0.033} & 0.008          & 0.013    \\ \hline
        Boston Housing Train & \textbf{0.000} & 1.000          & 0.001    \\ \hline
        Boston Housing Test  & \textbf{0.000} & 0.525          & 0.577    \\ \hline
        5-bit Parity         & \textbf{0.000} & \textbf{0.000} & 0.131    \\ \hline
        11-multiplexer       & \textbf{0.000} & \textbf{0.000} & 0.000    \\ \hline
        Santa Fe Ant Trail   & \textbf{0.000} & \textbf{0.000} & 0.317   \\ \hline
    \end{tabular}
\end{table}

\begin{table}[htbp]
\centering
\caption{\label{effectsize}Effect size between PSGE and GE, PGE and SGE.}
    \begin{tabular}{|l|c|c|c|}
    \hline
         \textbf{Problem}     & \textbf{PSGE-GE}     & \textbf{PSGE-PGE}    & \textbf{PSGE-SGE}   \\  \hline
        Quartic Polynomial   &  ++ &  + &  \texttildelow          \\  \hline
        Pagie Polynomial     &  +  &  -  &  -          \\  \hline
        Boston Housing Train & + &  \texttildelow &  -          \\  \hline
        Boston Housing Test  & ++ &  \texttildelow  &  \texttildelow         \\  \hline
        5-bit Parity         & +++ & +++ &  \texttildelow        \\  \hline
        11-multiplexer       & ++ & +++ &  - - -      \\  \hline
        Santa Fe Ant Trail   &  +  &  +  & \texttildelow  \\ \hline
    \end{tabular}
\end{table}

Focusing on the results of Tables \ref{statstable} and \ref{effectsize}, several observations about the approach proposed in this work can be drawn. In bold are the p-values of the comparisons where \gls{psge} is statistically better than \gls{ge}, \gls{pge} or \gls{sge}. 

\gls{psge} is statistically better than \gls{ge} on all problems. Comparing the performance of \gls{psge} with \gls{pge}, we see that \gls{psge} outperforms \gls{pge} with significant differences on four of the problems. In relation to the \gls{sge}, \gls{psge} never outperforms.

% ----------------------- QUAD
\begin{figure}[htbp]
	\centering
	\includegraphics[height=5cm]{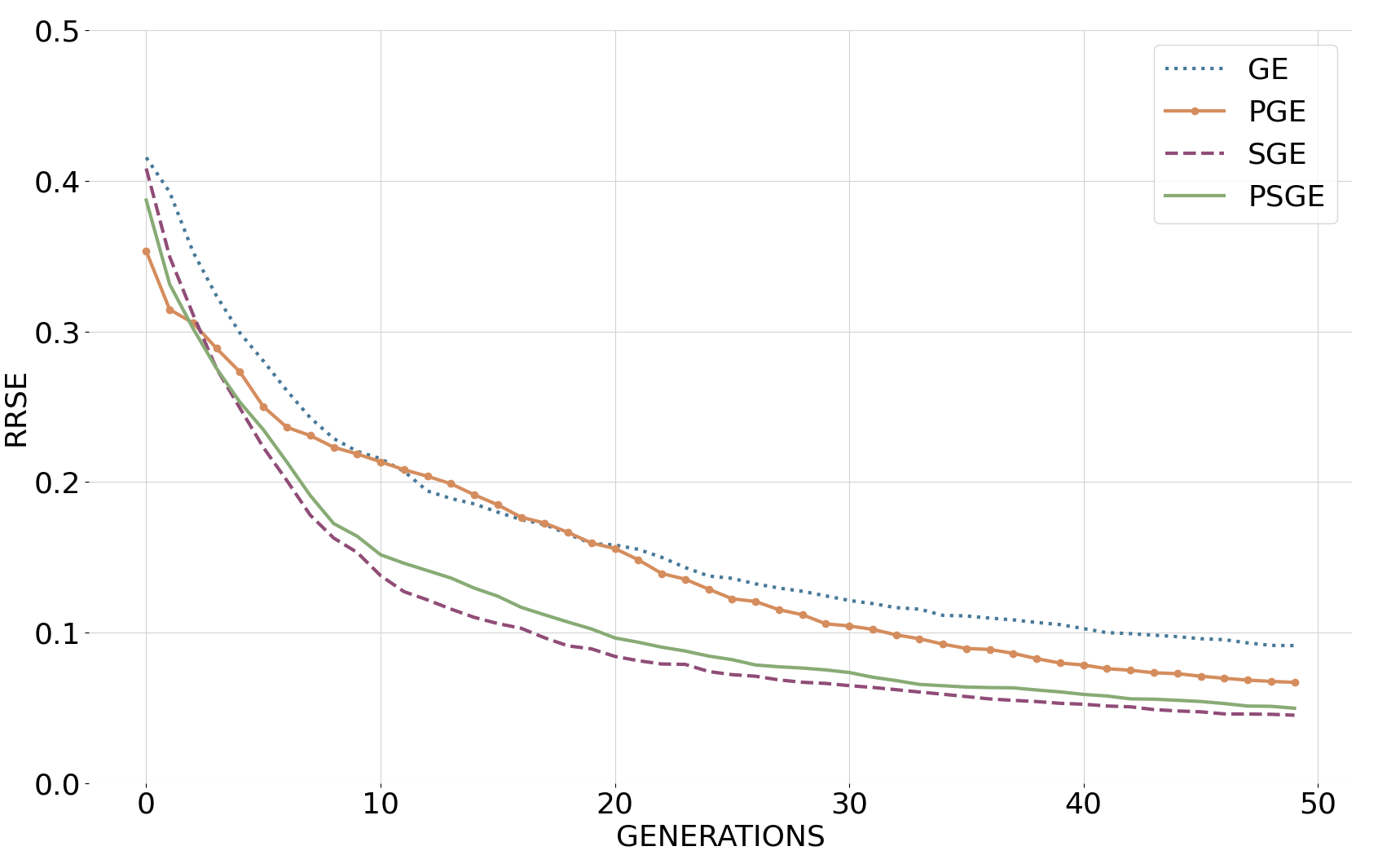}
	\caption{Performance results for the Quartic Polynomial. Results are the mean best fitness of 100 runs.}
	\label{fig:quadplot}
\end{figure}

Fig. \ref{fig:quadplot} shows the evolution of the mean best fitness for the quartic polynomial.
We can see that the decrease in \gls{sge} and \gls{psge} is similar, with \gls{sge} ending up with slightly better average fitness, but with no statistical differences between the methods (Table \ref{statstable}).
Looking at the \gls{pge} curve, we notice that it starts with better fitness, being surpassed by \gls{sge} and \gls{psge} around generation 3, decreasing more slowly in relation to the other methods. 
%From generation 20 the fitness starts to move away from the \gls{ge} values, however \gls{psge} is better statistically.
From generation 20 the fitness starts to move away from \gls{ge} values, however these never reach the performance of \gls{sge} and \gls{psge} during the 50 generations. \gls{psge} presents statistical differences from \gls{pge} with a small effect size.

% ------------- PAGIE
\begin{figure}[htbp]
	\centering
	\includegraphics[height=5cm]{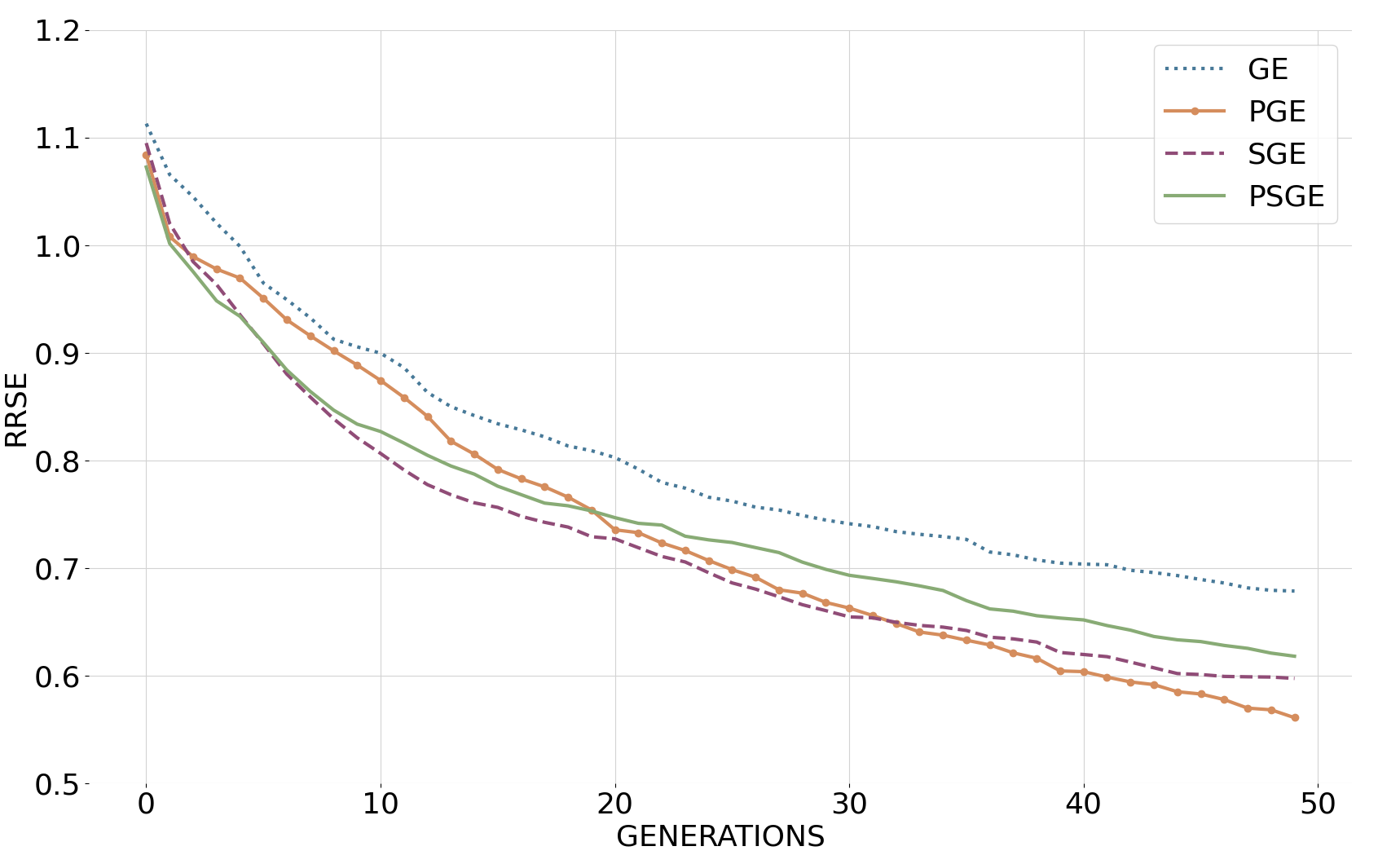}
	\caption{Performance results for the Pagie Polynomial. Results are the mean best fitness of 100 runs.}
	\label{fig:pagieplot}
\end{figure}

The results for the pagie polynomial are shown in Fig. \ref{fig:pagieplot}. By analysing the plot we see that all methods perform better than \gls{ge}, with the curves decreasing faster, and looking at the statistical test results in Table \ref{statstable} we see that \gls{psge} is better than \gls{ge} statistically with a small effect size. The method that ends up with the best average fitness is \gls{pge}, which is in line with the results presented in \cite{Megane2021}.%, and has no significant differences from \gls{psge}. %On the other hand, \gls{psge} presents statistical differences in relation to \gls{sge} (Table \ref{statstable}), and looking at the curve, it means that \gls{sge} has better performance in this problem. 

% ------------------- BOSTON HOUSING
% \begin{figure}
% 	\centering
% 	\includegraphics[height=5cm]{performance/bostonhousing.png}
% 	\caption{\label{subfig:bhplottrain}Performance results for Boston Housing - Training. Results are the mean best fitness of 100 runs.}
% \end{figure}

\begin{figure}[htbp]
	\centering
	\includegraphics[height=5cm]{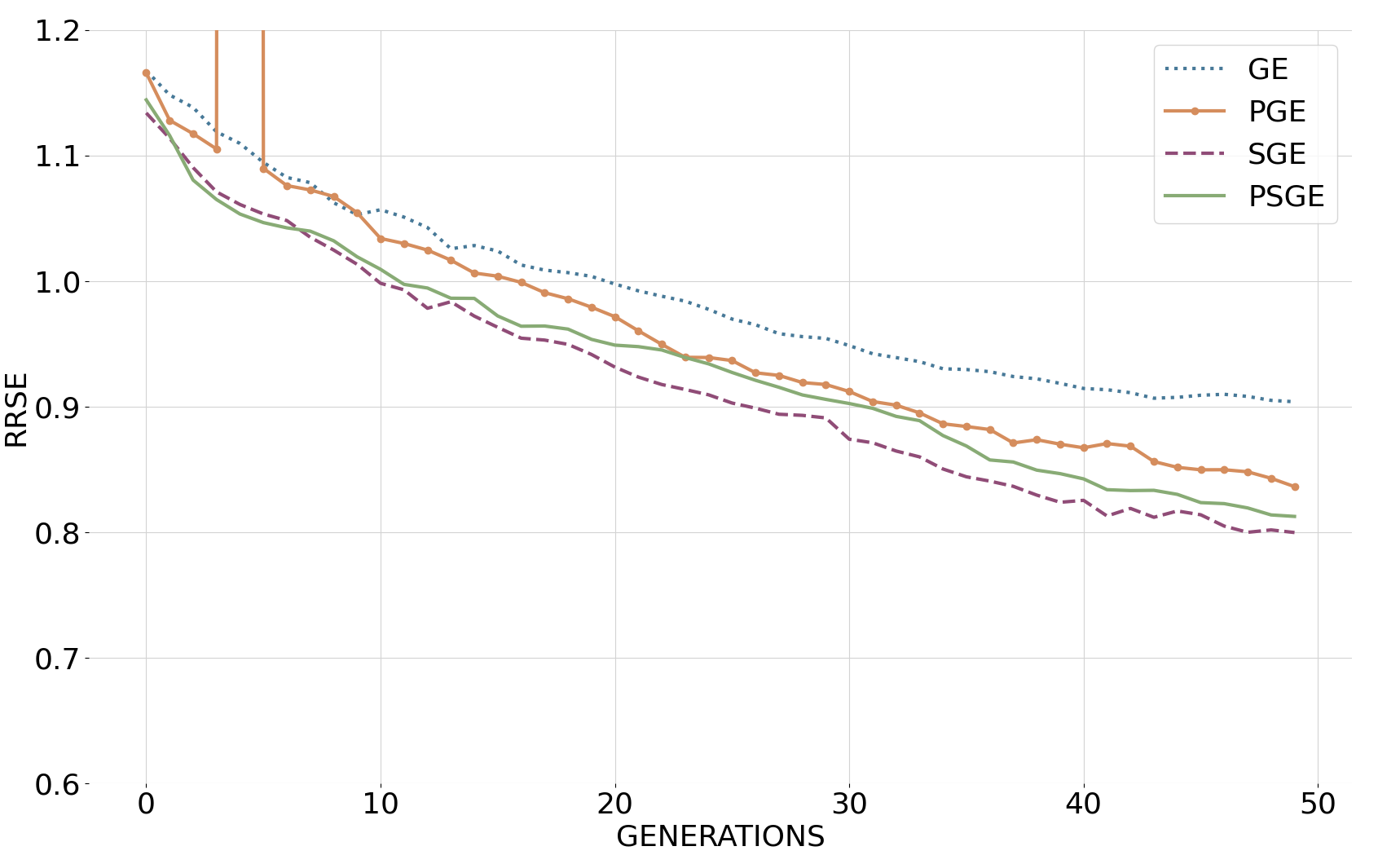}
	\caption{\label{subfig:bhplottest}Testing results for the Boston Housing. Results are the mean best fitness of 100 runs.}
\end{figure}

% ------------------------------------------- CORRIGIR; UM BOCADO PODRE ------------------------

Regarding the results for the Boston Housing problem, these are divided in 90\% for training and 10\% for testing.
The test results are the most relevant since they are the ones that evaluate the models with unseen data. The statistical tests show that in this problem \gls{psge} only presents statistical differences relative to \gls{ge}, having a small effect size in training and medium in testing.
\gls{psge} ends with better fitness than \gls{pge}, however these do not present significant differences.
We also see that when comparing with \gls{sge} in the training, \gls{psge} presents small statistical differences, being worse, however, in the test, these are not significant.
The small improvement in the results of test \gls{psge} over training when compared to \gls{ge} and \gls{psge}, may indicate that \gls{psge} has a better generalisation ability to predict unknown data.

% On Fig. \ref{subfig:bhplottrain} we can observe the training results for the Boston Housing problem, in which we observe that although \gls{psge} has a slightly faster growth, at the end of 50 generations \gls{pge} reaches the average fitness obtained by \gls{psge}. The statistical tests show that \gls{pge} and \gls{psge} are statistically similar in this problem. The curve of \gls{ge} decreases more slowly, with the fitness ending in worse values, making \gls{psge} better statistically. \gls{sge} is the method that shows the greatest decrease, and it is statistically better than \gls{psge}.

% Regarding the test results (Fig. \ref{subfig:bhplottest}), we see that the \gls{psge} results are close to the \gls{sge}, and there are no statistical differences between the methods. The test results are usually the most relevant for evaluating the trained models since they allow observing their behaviour with unseen data.

% ----------------- 5 parity

\begin{figure}[htbp]
	\centering
	\includegraphics[height=5cm]{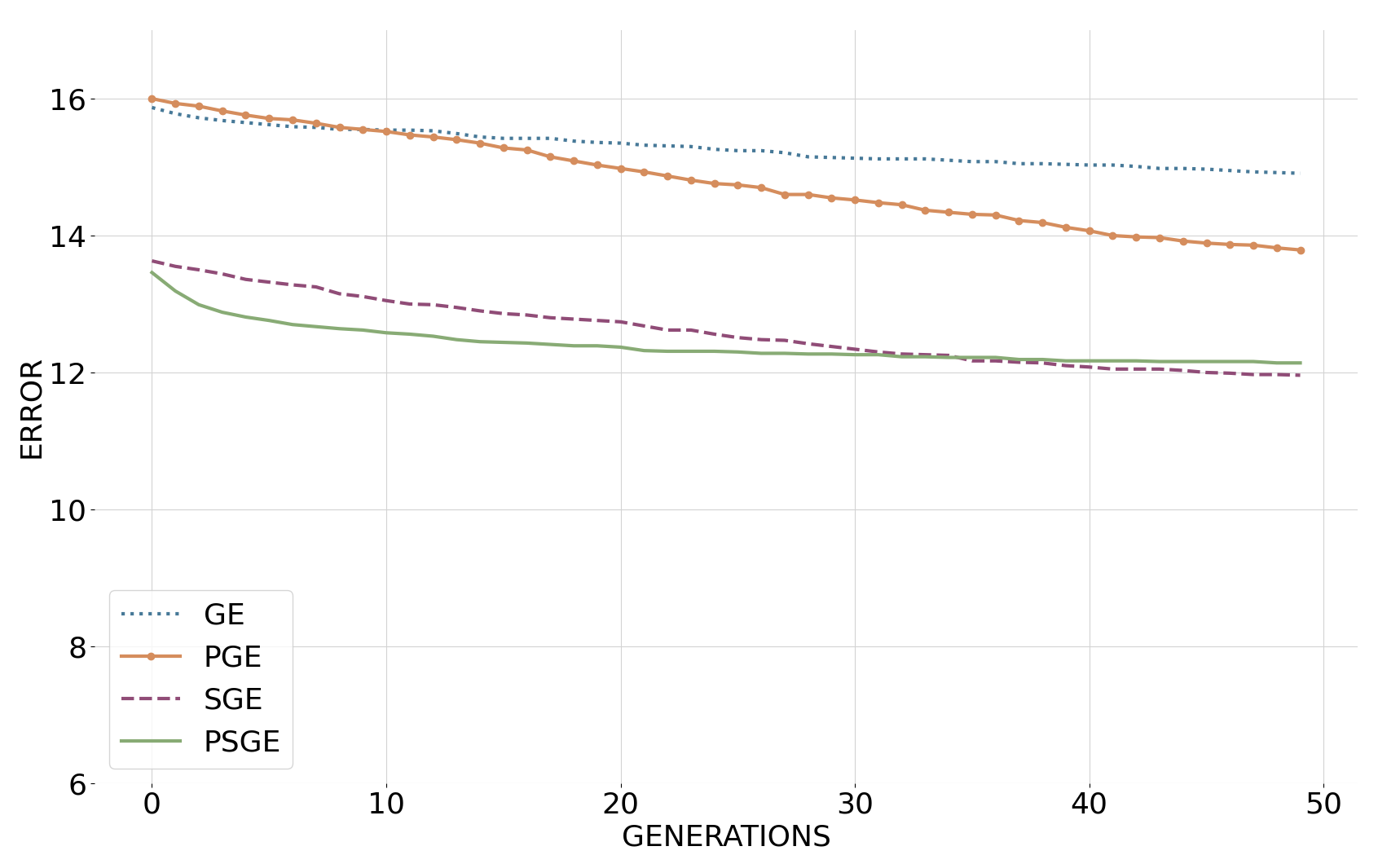}
	\caption{Performance results for the 5-bit Even Parity problem. Results are the mean best fitness of 100 runs.}
	\label{fig:5parplot}
\end{figure}

Looking at the results for the 5-bit Even Parity problem, we observe that the performance of \gls{sge} and \gls{psge} is better than that of \gls{ge} and \gls{pge}, starting with lower fitness values. According to the statistical results, \gls{psge} is statistically better than \gls{ge} and \gls{pge}, with a large effect size. In comparison to \gls{sge} we see that the decrease of \gls{psge} is slightly faster, maintaining better average fitness for 30 generations, and then it is reached by \gls{sge}, presenting no significant differences between the two methods.

\begin{figure}[htbp]
	\centering
	\includegraphics[height=5cm]{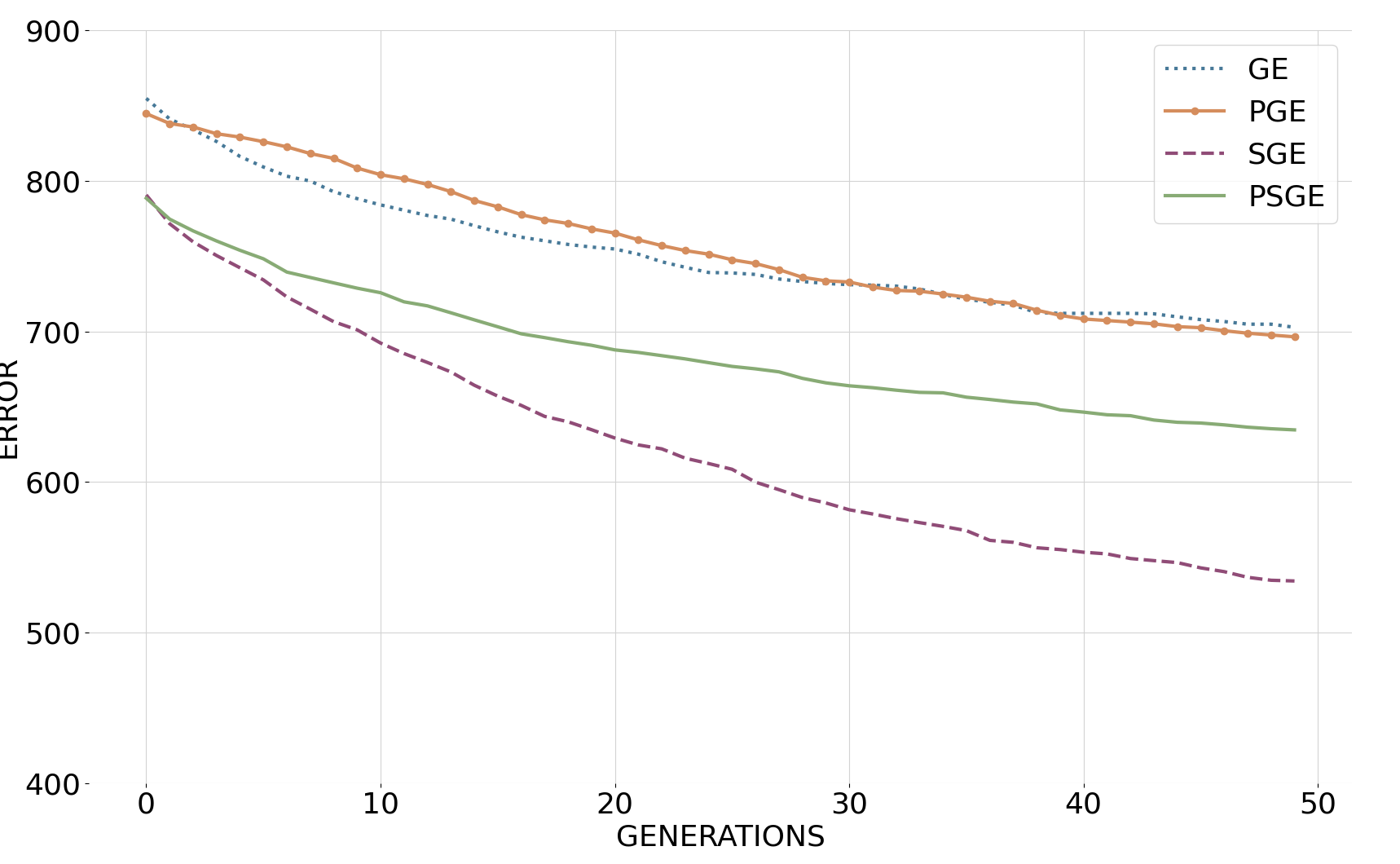}
	\caption{Performance results for the 11-bit Boolean Multiplexer problem. Results are the mean best fitness of 100 runs.}
	\label{fig:11multplot}
\end{figure}

Regarding the results of the tests done for the 11-bit Boolean Multiplexer (Fig. \ref{fig:11multplot}), we can see that the methods behave quite differently. As observed previously for the 5-bit parity problem (Fig. \ref{fig:5parplot}), the average fitness of \gls{sge} and \gls{psge} individuals at the beginning of the evolutionary process is better, with significant differences between \gls{psge} and \gls{ge}, with a medium effect size, and also with \gls{pge} presenting a large effect size. In this problem we see that the decrease of \gls{sge} is much steeper, rapidly moving away from the average fitness of \gls{psge}.

% ---------------- Santa Fe ant trail

\begin{figure}[htbp]
	\centering
	\includegraphics[height=5cm]{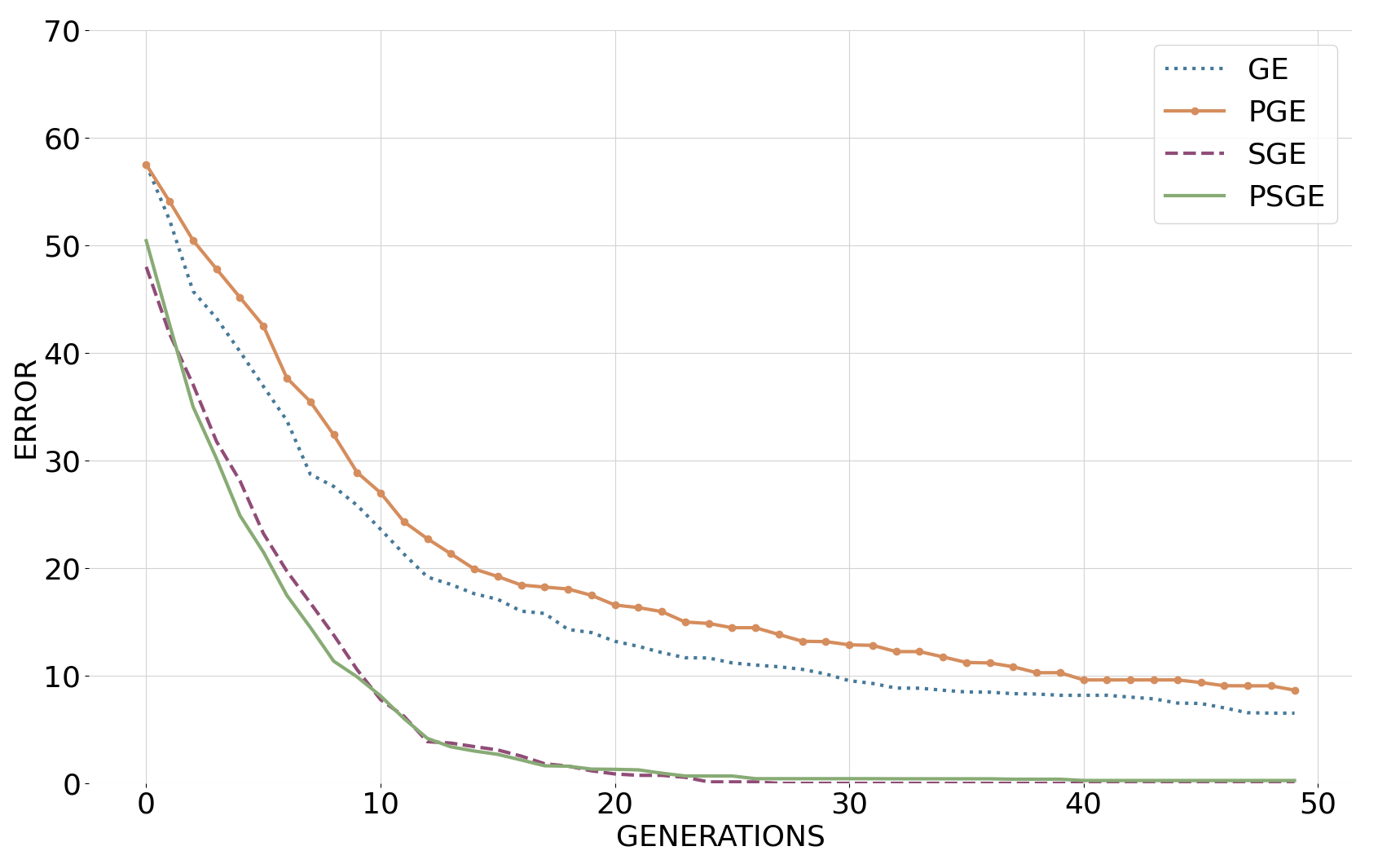}
	\caption{Performance results for the Santa Fe Artificial Ant problem. Results are the mean best fitness of 100 runs.}
	\label{fig:antplot}
\end{figure}

The last analysis studies the performance on the Santa Fe Ant problem (Fig. \ref{fig:antplot}). In contrast to the previous analyses, in this problem the method that performs worst over generations is \gls{pge}. On the other hand we observe that the fitness of the individuals of \gls{sge} and \gls{psge} decreases rapidly, approaching the optimal solution after 20 generations. \gls{psge} is statistically better than \gls{ge} and \gls{pge}, not presenting statistical differences in relation to \gls{sge}.

\section{Conclusion}
\label{chap:conclusion}

% TODO: falar que usar uma PCFG atualizada tambem pode ser util em problemas do mundo real, com o conhecimento que temos do problema , para guiar a procura
% falar em algum lado de gramaticas balanceadas e de gramaticas explosivas - seria interessante dizer no future work , analisar o comportamento do algoritmo com diferentes gramaticas. dizer algo tipo "na tese fizemos uma pequena analise do comportamento dos algoritmos com gramaticas já evoluidas, no entanto é tambem importante analisar o comportamento destas com diferentes gramaticas"

% falar tambem que a analise da evoluçao das 
\gls{ge} is a grammar-based \gls{gp} variant that has attracted the attention of many researchers and practitioners, since its proposal in the late 1990s and it has been applied with success to many problem domains. However, it has been shown that it suffers from some issues, such as low locality and high redundancy \cite{Rothlauf2003,Rothlauf2006}.  

In grammar-based \gls{gp}, the choice of the grammar has a significant impact on the quality of the generated solutions as it is the grammar that defines the space of possible solutions. Our goal work was to create a variant of \gls{ge} that could guide the search towards better solutions, inserting bias in the production rules that generate better individuals, in order to achieve a better overall performance, and at the same time overcome the problems of \gls{ge}.

In this paper we present \gls{psge}, a new variant of \gls{ge} that introduces a new representation alternative to \gls{sge}, using \gls{pge}'s mapping mechanism. The genotype is a set of dynamic lists, one for each non-terminal of the grammar. Each codon in the list represents the likelihood of selecting a production rule. At the end of each generation the \gls{pcfg} is updated and individuals are remapped, same as in \gls{pge}.

The proposed method is compared to the standard versions of \gls{ge}, \gls{pge} and \gls{sge} on different benchmark problems, analysing the evolution of the mean best fitness.
\gls{psge} outperformed \gls{ge} with statistical differences in all problems, while it outperformed \gls{pge} in 4 of the 6 problems. Regarding the relative performance with \gls{sge}, we show that the methods obtain similar performances.

\gls{psge} proved to be a good alternative to \gls{pge}, since it presents better performance, keeping the advantage of using a probabilistic grammar.
At the end of the evolutionary process we get a specialised grammar for the problem at hand which can be used to obtain information about which production rules are more relevant to create the best individuals. The use of a \gls{pcfg} can also be used by manually guide the search, for example, in the real-world problems where we have some information regarding some problems, we can alter the probabilities of the grammar by introducing some known biases.
As future work it will be interesting to analyze the average fitness of sample populations created with a previously evolved grammar and also to analyze the fitness evolution over multiple generations.
Another line of work that will be interesting to analyze is to test the algorithm behavior with the probabilities of grammars initialized randomly, or with different types of grammars, similar to the study done by Nicolau et al. \cite{Nicolau2018}.

\section*{Acknowledgment}
This work was partially funded by the project grant DSAIPA/DS/0022/2018 (GADgET), by FEDER funds through the Operational Programme Competitiveness Factors - COMPETE and national funds by FCT - Foundation for Science and Technology (POCI-01-0145-FEDER-029297, CISUC - UID/CEC/ 00326/2020) and within the scope of the project A4A: Audiology for All (CENTRO-01-0247-FEDER-047083) financed by the Operational Program for Competitiveness and Internationalisation of PORTUGAL 2020 through the European Regional Development Fund.

% The preferred spelling of the word ``acknowledgment'' in America is without 
% an ``e'' after the ``g''. Avoid the stilted expression ``one of us (R. B. 
% G.) thanks $\ldots$''. Instead, try ``R. B. G. thanks$\ldots$''. Put sponsor 
% acknowledgments in the unnumbered footnote on the first page.

% \section*{References}

% Please number citations consecutively within brackets \cite{b1}. The 
% sentence punctuation follows the bracket \cite{b2}. Refer simply to the reference 
% number, as in \cite{b3}---do not use ``Ref. \cite{b3}'' or ``reference \cite{b3}'' except at 
% the beginning of a sentence: ``Reference \cite{b3} was the first $\ldots$''

% Number footnotes separately in superscripts. Place the actual footnote at 
% the bottom of the column in which it was cited. Do not put footnotes in the 
% abstract or reference list. Use letters for table footnotes.

% Unless there are six authors or more give all authors' names; do not use 
% ``et al.''. Papers that have not been published, even if they have been 
% submitted for publication, should be cited as ``unpublished'' \cite{b4}. Papers 
% that have been accepted for publication should be cited as ``in press'' \cite{b5}. 
% Capitalize only the first word in a paper title, except for proper nouns and 
% element symbols.

% For papers published in translation journals, please give the English 
% citation first, followed by the original foreign-language citation \cite{b6}.
\bibliographystyle{IEEEtran}
\bibliography{IEEEabrv,bibliography.bib}

\end{document}